\documentclass[11pt]{article}

\usepackage[preprint]{acl}

\usepackage{times}
\usepackage{latexsym}
\usepackage{subcaption}
\usepackage{stfloats}
\usepackage{tikz}
\usepackage{tcolorbox}
\tcbuselibrary{skins}
\usepackage{multirow} 
\usepackage{makecell}
\usepackage{array}
\usepackage[table]{xcolor}

\usepackage[T1]{fontenc}

\usepackage[utf8]{inputenc}

\usepackage{microtype}

\usepackage{inconsolata}

\usepackage{graphicx}
\usepackage{booktabs}
\usepackage{amsfonts}

\title{iLENS: \underline{I}nterpretable \underline{L}LM-Guided Mixture-of-\underline{E}xperts \\ 
for \underline{N}euroimaging \underline{S}urvival Analysis}

\author{
\textbf{Farica Zhuang\textsuperscript{1}},
 \textbf{Seong Woo Han\textsuperscript{1}},
 \textbf{Zixuan Wen\textsuperscript{1}},
 \textbf{Shu Yang\textsuperscript{1}}, \\
 \textbf{Yize Zhao\textsuperscript{2}},
 \textbf{Li Shen\textsuperscript{1}}, \\
  \textsuperscript{1}University of Pennsylvania,
 \textsuperscript{2}Yale University
 }

\begin{document}
\maketitle
\begin{abstract}
Alzheimer's Disease (AD) is a complex neurodegenerative disorder that continues to impact millions of people worldwide. Predicting AD conversion during the prodromal stage remains critical for disease understanding and patient care. As such, survival models are widely used for AD risk prediction, yet they are typically static predictors with limited interpretability and no capacity for natural language reasoning. In this work, we propose iLENS, an interpretable large language model (LLM) guided framework based on mixture-of-experts (MoE) for survival prediction in AD conversion. Our approach uses LLM to synthesize structured neuroimaging measurements and unstructured information to guide expert routing. Our framework demonstrates competitive predictive performance and capability in patient subtyping. Furthermore, our framework provides transparent, biologically grounded rationales for its routing decisions, bridging the gap between high-performance survival analysis and interpretable clinical decision support. 

\end{abstract}

\section{Introduction}

Survival analysis, or time-to-event prediction, is an important component of precision medicine. In neurodegenerative disorders, such as Alzheimer's Disease (AD), predicting risk scores for conversion to AD and identifying specific risk profiles are essential for early intervention and resource allocation \cite{sperling2011toward, nakagawa2020prediction, mirabnahrazam2023predicting}. Classical survival methods and modern deep learning models have improved predictive accuracy, yet they remain primarily as predictors for numerical outcome risk scores \cite{fox2002cox-ph, nagpal2021dsm, katzman2018deepsurv}. In clinical settings, however, the ability to identify patient subgroups or clusters based on risk profiles is valuable, as such subtypes can reveal heterogeneous disease progression patterns and support more interpretable prognostic decision-making. This is particularly important in AD, where patients often exhibit diverse trajectories of neurodegeneration and biomarker burden \cite{goyal2018characterizing}. To bridge this gap, we introduce iLENS, an interpretable LLM-guided MoE for neuroimaging survival analysis. Moving beyond static encoders, we investigate whether language models can serve as semantic  routing controllers within an expert-based survival clustering framework. This design provides two-layer interpretability. First, we provide natural language interpretability based on each patient's structured neuroimaging biomarker measurements to justify the model's clinically-grounded expert routing. Second, we maintain clustering interpretability and uncover the underlying survival subtypes. We evaluate our framework using real-world longitudinal neuroimaging data from the Alzheimer's Disease Neuroimaging Initiative (ADNI) dataset \cite{petersen2010alzheimer} to predict and subtype AD conversion. 





\begin{figure*}[t]
  \centering
  \includegraphics[width=0.75\linewidth]{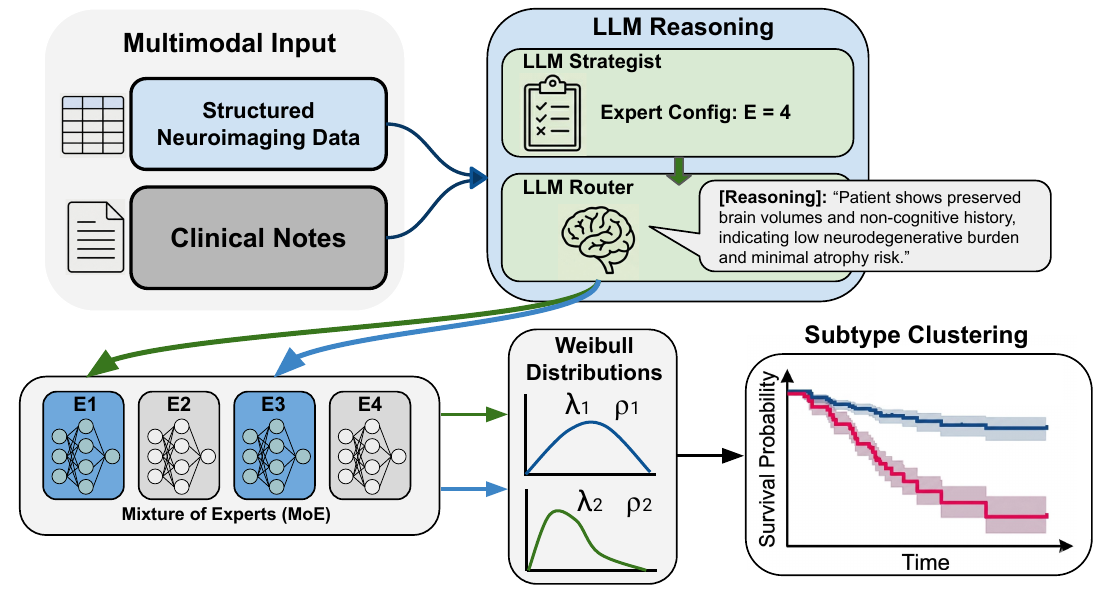}
  \caption{\textbf{Overview of iLENS.} The system processes multimodal inputs of structured MRI features and clinical notes. An LLM Strategist defines the routing configuration, while an LLM Router assigns weights to specialized experts within a Mixture-of-Experts bank. This process generates an interpretable clinical reasoning rationale and a mixed Weibull distribution to produce individualized survival analysis curves for Alzheimer's disease progression.}
  \label{fig:schematics}
\end{figure*}

\section{Related Work}

\textbf{Survival Prediction and Subtype Clustering.} Traditional survival models often focus on risk score predictions for a given outcome \cite{fox2002cox-ph, katzman2018deepsurv, nagpal2021dsm}. However, a crucial goal of survival prediction in medical applications is also to identify patient
subgroups with similar survival patterns and risk profiles, known as subtyping (e.g., high risk v.s. low risk) \cite{abbasi2024survival}. This helps with treatment assignment, resource allocation, and disease understanding \cite{carobbio2020multistate, glare2003systematic, binder2008allowing}. Hence, neural survival clustering (NSC) models were introduced as a new class of survival models that not only predict individualized risk, but also perform explicit subtype discovery \cite{jeanselme2022neural-nsc, hou2023dcsm-ieee, hou2024DCSM, wen2025multimodel-dcsm}. Yet, these models rely on a single shared encoder, which often fails to capture distinct feature patterns in heterogeneous populations and disease structures, commonly seen in complex diseases, where AD patient profiles vary in atrophy and other biomarkers. 

\textbf{MoE in Clinical Modeling.} MoE architectures demonstrate improved ability to capture heterogeneous data distribution through specialized sub-networks, known as experts \cite{fedus2022switch, zhu2024unimed, shen2024mome}. The routing logic of the experts is learned by a gating network, typically a neural network. Standalone MoE architectures have since been applied to clinical tasks, showing strengths in handling data heterogeneity, interpretability, and missing modalities in clinical scenarios \cite{zhang2025surmoe, yun2024flex, xin2025i2moe, zhuang2025mref}. However, MoE interpretability is typically limited to the expert weights from the gating network, leaving the routing logic needing to be analyzed post-hoc and reducing transparency in clinical setting.

\textbf{LLM for Clinical Decision Support.} LLM is increasingly utilized to perform various tasks, plan, and make predictions in single- or multi-agent frameworks for clinical tasks \cite{tang2024medagents, wang2025medagent-pro, hou2025adagent, li2024mmedagent, kim2024mdagents}. Given these capabilities, recent advances further demonstrate that LLMs can serve as a router, acting as the initial processor of information before making a decision where a task should go \cite{liu2025llmoe, ong2024routellm, hu2024routerbench}. Unlike traditional classifiers, LLM routers can ingest natural language instructions and reasoning to improve routing decisions beyond structured data alone in clinical applications.

\section{Method}

\subsection{LLM-Guided Routing}
Unlike traditional MoE models that rely on a gating network, our framework utilizes the LLM semantic reasoning to define and guide the expert specialization process in two stages as shown in Figure~\ref{fig:schematics}. We use GPT 4.1-mini as the LLM for the two stages.

\textbf{Phenotype-Driven Expert Initialization.} We utilize an LLM to map the high-dimensional structured feature space from the list of neuroimaging measurement regions and the list of unstructured clinical notes into a set of $E$ distinct physiological phenotypes (Appendix~\ref{appendix:expert_design}, Table~\ref{tab:av45_prompt}). Given the set of feature names $F$ and the prognostic task, the LLM generates an architectural specification $S = \{e_{1},e_{2},...,e_{E}\}$, where each expert definition $e_i$ contains information such as the expert name and description (Table~\ref{tab:av45_experts}). In our setting, we use $E=4$. By defining $E$ with the prompt, we ensure the MoE backbone is anchored in a fixed number of clinically separable AD progression patterns.

\textbf{Semantic Routing.} For a given patient with structured neuroimaging features $x_{neuro}$ (i.e., measured brain regions of interest) and unstructured clinical notes $x_{notes}$ (i.e., initial assessment and medical history), the routing weights $g(x)$ are assigned by the LLM based on the patient's alignment with the initialized phenotypes $S$. ROI features are aggregated into mesoscale brain region summaries to improve semantic interpretability (Appendix~\ref{appendix:mesoscale}, Table~\ref{tab:mesoscale_regions}). The LLM serves as the semantic gating function:
\begin{equation}
    g(x) = \mbox{Softmax}(\mbox{LLM}(x_{neuro}, x_{notes},S)). 
\end{equation}
An example of the prompt and routing output are in Appendix~\ref{appendix:llm-routing}, Table~\ref{tab:av45_llm_router_prompt}. The LLM outputs a probability distribution and clinical reasoning for the top $k$ experts, where we define $k=2$ to mimic sparsity in MoE. The reasoning provides human-readable interpretation of the routing decisions.

\begin{table*}[t]

\caption{\textbf{Performance Comparison.} Comparison across brain imaging modalities (VBM, FDG, AV45) and cluster counts ($K$). iLENS demonstrates improved or competitive prognosis (C-index) and subtype separation (LogRank) across nearly all settings while maintaining competitive prognostic discrimination. For each metric, the best-performing model is highlighted in \textbf{bold} and the second-best is \underline{underlined}. $-$: LogRank undefined due to assignment of test samples to a single cluster.}
\label{tab:multimodal_results}

\centering
\small

\begin{tabular}{c c | cc | cc | cc}




\toprule
\multirow{2}{*}{\makecell{\textbf{Cluster}\\\textbf{Count}}}
& \multirow{2}{*}{\textbf{Model}}
& \multicolumn{2}{c|}{\textbf{VBM (MRI)}} 
& \multicolumn{2}{c|}{\textbf{FDG (Metabolism)}} 
& \multicolumn{2}{c}{\textbf{AV45 (Amyloid)}} \\

& 
& C-Idx $\uparrow$
& Log-R $\uparrow$
& C-Idx $\uparrow$
& Log-R $\uparrow$
& C-Idx $\uparrow$
& Log-R $\uparrow$ \\
\midrule

\multirow{5}{*}{$K$=2}
& SCA & 0.502 & 1.63 & 0.588 & 1.78 & 0.561 & 0.79 \\

& VaDeSC & 0.548 & 2.45 & 0.525 & $-$ & 0.618 & \underline{28.73} \\

& NSC & 0.604 & \textbf{4.25} & \underline{0.679} & \underline{6.56} & 0.728 & $-$ \\

& DCSM & \underline{0.638} & $-$ & 0.654 & $-$ & \textbf{0.786} & 16.81 \\

& \cellcolor{gray!15}\textbf{iLENS (Ours)}

& \cellcolor{gray!15}\textbf{0.670}

& \cellcolor{gray!15}\underline{2.92}

& \cellcolor{gray!15}\textbf{0.775}

& \cellcolor{gray!15}\textbf{11.22}

& \cellcolor{gray!15}\underline{0.759}

& \cellcolor{gray!15}\textbf{36.34} \\

\midrule

\multirow{5}{*}{$K$=3}
& SCA & 0.440 & 0.60 & 0.450 & 3.96 & 0.527 & 4.46 \\

& VaDeSC & 0.483 & 1.55 & 0.562 & 0.74 & 0.565 & 14.81 \\

& NSC & 0.529 & 0.72 & 0.690 & \underline{15.50} & 0.508 & 3.49 \\

& DCSM & \underline{0.595} & \textbf{7.07} & \textbf{0.803} & 9.31 & \textbf{0.783} & \underline{20.24} \\

& \cellcolor{gray!15}\textbf{iLENS (Ours)}

& \cellcolor{gray!15}\textbf{0.634}

& \cellcolor{gray!15}\underline{2.54}

& \cellcolor{gray!15}\underline{0.722}

& \cellcolor{gray!15}\textbf{25.97}

& \cellcolor{gray!15}\underline{0.767}

& \cellcolor{gray!15}\textbf{29.01} \\

\bottomrule
\end{tabular}

\end{table*}

\subsection{Expert-Routed Survival Clustering}
Given the expert routing weights $g(x) \in \mathbb{R}^E$, each patient representation $h(x)$ is obtained as a weighted combination of expert outputs, 
\begin{equation}
    h(x) = \sum_{e=1}^{E} g_e(x) f_e(x),
\end{equation}
where $f_e(x)$ denotes the output of expert $e$ and $g_e(x)$ is its routing weight. To identify clinically meaningful subgroups, we model the survival process as a latent mixture of $K$ distinct survival subtypes. Following the generative approach of previous clustering survival models such as DCSM \cite{hou2024DCSM}, the survival function for a patient with representation $h(\mathbf{x})$ is defined as 
\begin{equation}
    S(t \mid \mathbf{x}) = \sum_{k=1}^{K} \pi_k(\mathbf{x}) S_k(t),
\end{equation}
where $\pi_k(\mathbf{x})$ represents the probability of the patient belonging to the $k$-th survival subtype. In our implementation, we set $K=2$ corresponding to a high-risk and low-risk stratification. Each subtype survival function $S_k(t)$ is modeled using a Weibull distribution, chosen for its flexibility and closed-form expression, 
\begin{equation}
S_k(t) = \exp\left({-\left(\frac{t}{\lambda_k}\right)^{\rho_k}}\right),
\end{equation}
where $\lambda_k$ and $\rho_k$ denote the scale and shape parameters, respectively. Unlike traditional models that estimate these parameters per-patient, we treat them as global parameters that define the shared survival trajectory of each latent subtype. The assignment weights $\pi_k(\mathbf{x})$ are estimated from the LLM-guided expert-enhanced representation $h(\mathbf{x})$ via a softmax layer:
\begin{equation}
\pi_k(\mathbf{x}) = \frac{\exp(w_k^T h(\mathbf{x}))}{\sum_{j=1}^{K} \exp(w_j^T h(\mathbf{x}))}.
\end{equation}
These weights act as soft cluster memberships, allowing the model to provide an individualized survival prediction while maintaining interpretablility to population-level risk profiles. Details of the LLM-guided survival modeling framework are provided in Appendix~\ref{appendix:survival-model-after-llm-routing}.

\section{Experiments}

\subsection{Dataset}

We evaluate iLENS on the Alzheimer's Disease Neuroimaging Initiative (ADNI) \cite{petersen2010alzheimer} dataset, a longitudinal multimodal study that includes clinical, imaging, genetic, and biochemical biomarkers. In this work, we obtained three structured imaging measurements from ADNI, namely the VBM, FDG, and AV45 for brain volume and biomarker measurements. For clinical notes, we extracted the initial assessment and recent medical history (Appendix~\ref{appendix:dataset}).

\begin{figure*}[!t]
\centering

\begin{subfigure}{0.32\textwidth}
    \centering
    \includegraphics[width=\linewidth]{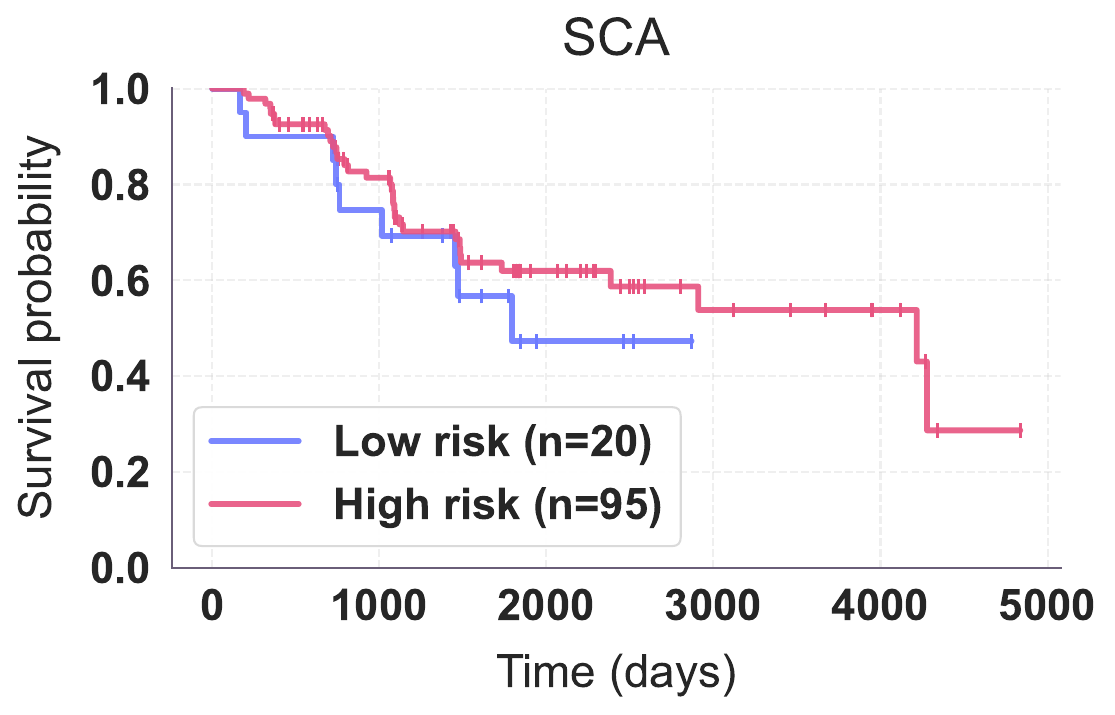}
\end{subfigure}
\hfill
\begin{subfigure}{0.32\textwidth}
    \centering
    \includegraphics[width=\linewidth]{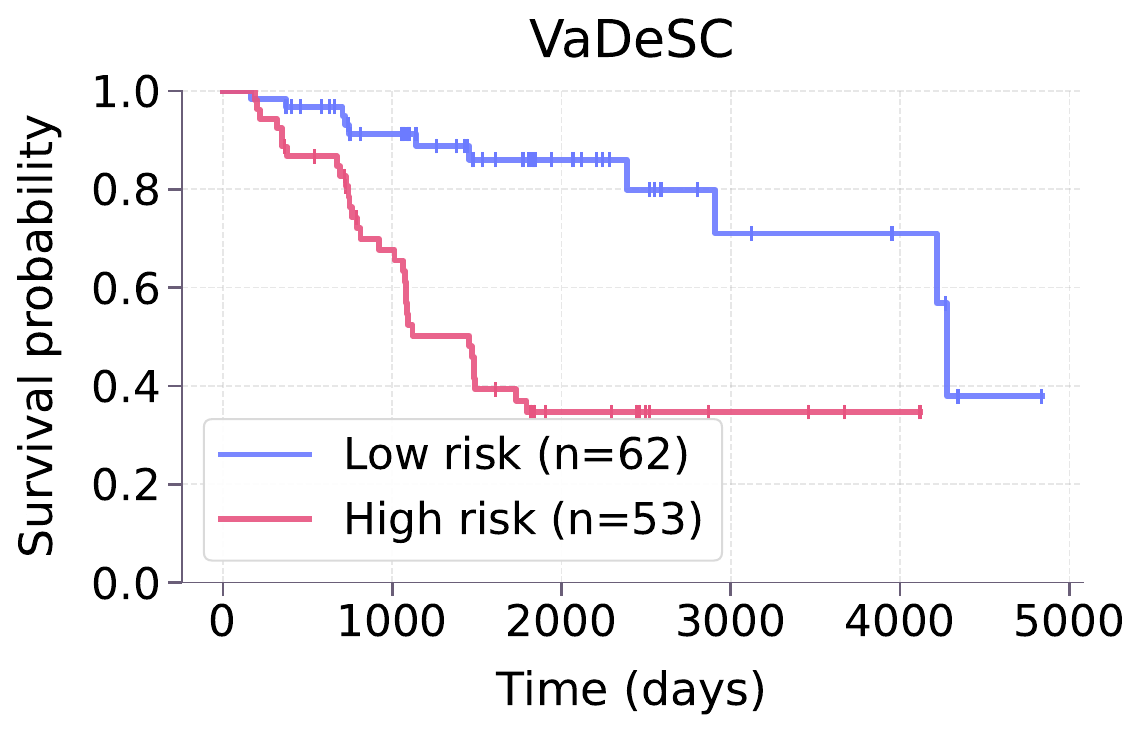}
\end{subfigure}
\hfill
\begin{subfigure}{0.32\textwidth}
    \centering
    \includegraphics[width=\linewidth]{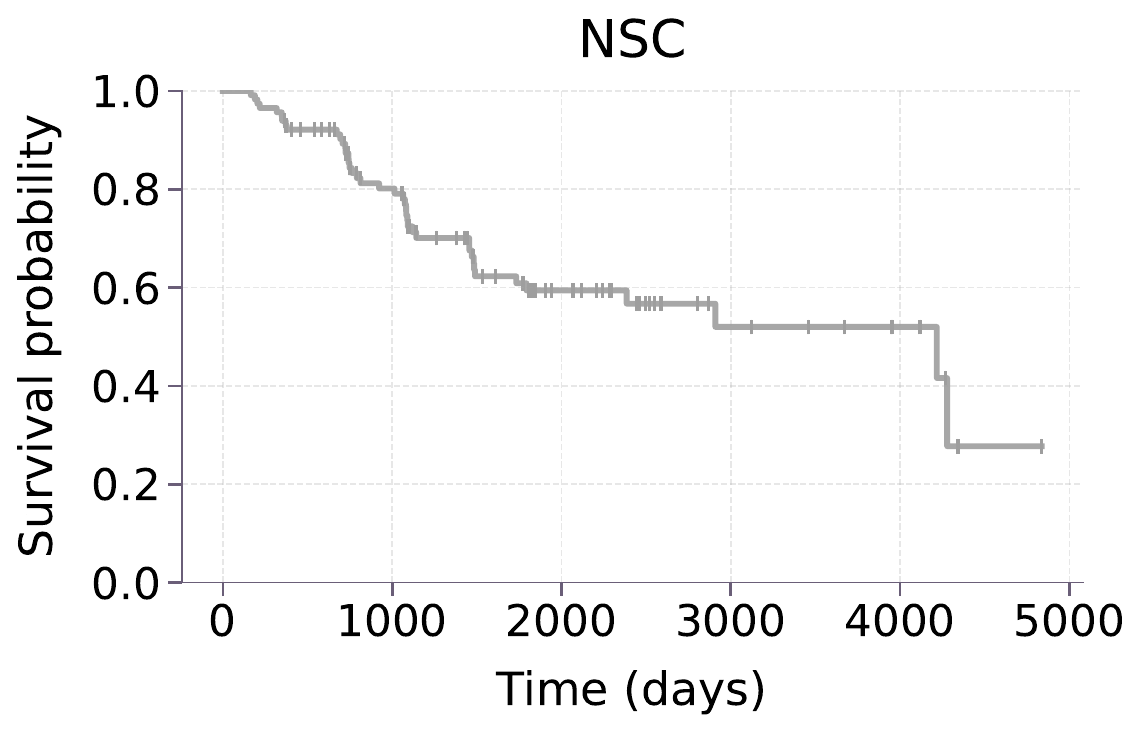}
\end{subfigure}

\vspace{0.5em}

\makebox[\textwidth][c]{

\begin{subfigure}{0.32\textwidth}

    \centering

    \includegraphics[width=\linewidth]{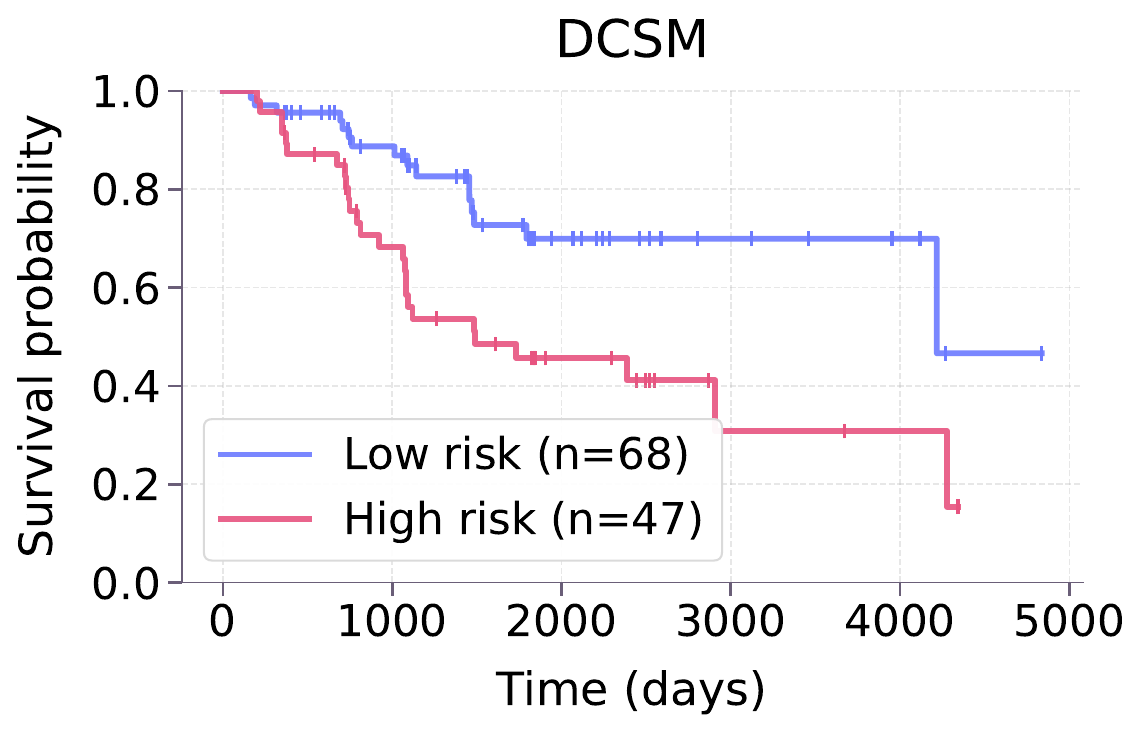}

\end{subfigure}

\hspace{1em}

\begin{subfigure}{0.32\textwidth}

    \centering

    \includegraphics[width=\linewidth]{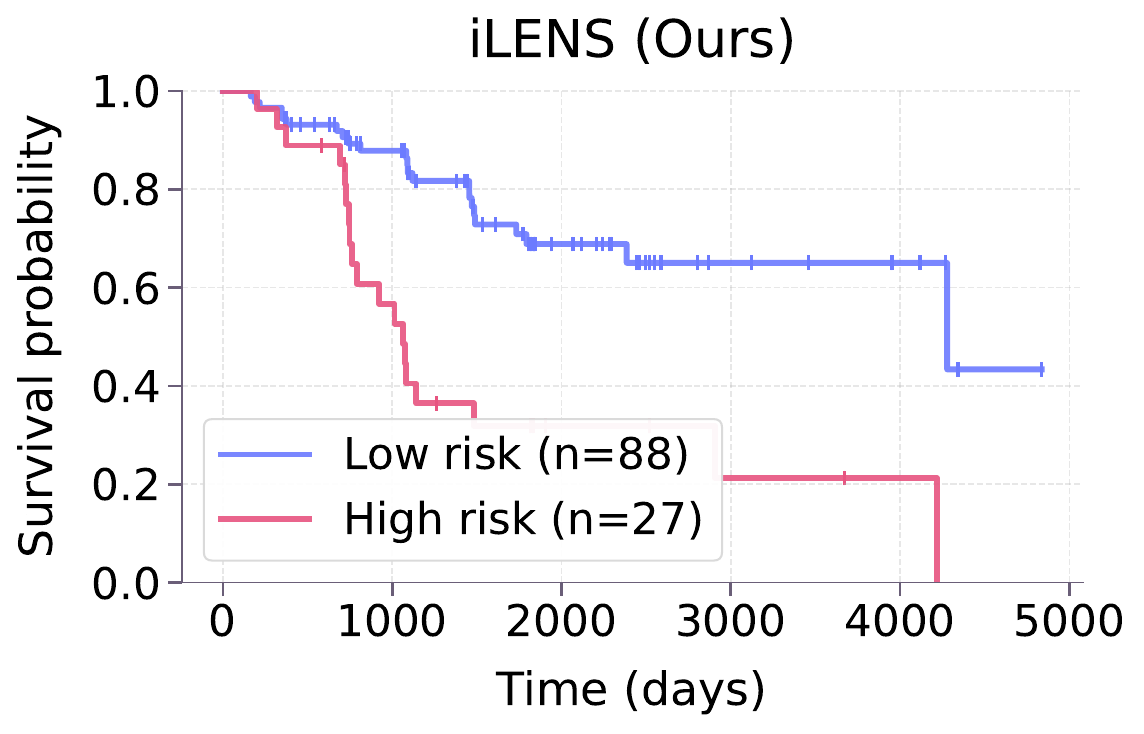}

\end{subfigure}

}

\caption{\textbf{Subtype Separation.} Kaplan--Meier survival curves for AV45 with $K=2$ across clustering methods. Blue and red curves indicate lower- and higher-risk subgroups respectively according to the mean predicted survival risk. Gray curve indicates that the test samples collapsed to a single cluster.}
\label{fig:km-av45-k2}
\end{figure*}

\subsection{Baselines}
We compare iLENS with state-of-the-art clustering survival models, including SCA \cite{chapfuwa2020survival-sca}, VaDeSC \cite{manduchi2021deep-vadesc}, NSC \cite{jeanselme2022neural-nsc}, and  
 Deep Clustering Survival Machine (DCSM) \cite{hou2024DCSM}. All models are hyperparameter-tuned using Optuna \cite{akiba2019optuna} (Appendix~\ref{appendix:training_and_eval}, Table~\ref{tab:hyperparameters}).

\subsection{Interpretable Prognosis and Subtyping}

We first compare our framework against baselines using Concordance Index (C-index) for prognostic performance and LogRank statistic to measure separation between discovered subtypes. In practice, there is often a competitive trade-off between individual prognostic accuracy and population-level subtype separation, as seen in other baselines commonly sacrificing one for the other. However, our results show that iLENS maintains competitive performance for both C-index and LogRank (Table~\ref{tab:multimodal_results}) across modalities and cluster counts, suggesting that LLM-guided expert routing better captures heterogeneous AD progression patterns compared to a shared encoder. Figure \ref{fig:km-av45-k2} illustrates that iLENS produces more distinct survival subtype clusters compared to all baselines, reflecting clinically differentiable risk trajectories (Appendix~\ref{appendix:km_plots_av45_2_clusters}) Importantly, iLENS also produces direct natural language rationale for expert assignments, which are grounded in each patient's brain imaging measurements and medical history (Table~\ref{tab:semantic_routing_example}). We provide qualitative examples of interpretable semantic routing in Appendix~\ref{appendix:interpretability}. Additionally, to show the effectiveness of semantic LLM-guided routing, we conduct ablation studies against non-semantic and non-specialized routing strategies, where we see that sparse LLM routing outperforms other sparse routing settings (Appendix~\ref{appendix:ablations}, 
Tables~\ref{tab:ablation_av45}, 
\ref{tab:ablation_fdg}, and~\ref{tab:ablation_vbm}).

\section{Conclusion}
We introduced iLENS, a new LLM-guided MoE framework for survival prediction. By integrating the semantic reasoning of LLM with the capability of MoE in handling inherently heterogeneous clinical data, iLENS offers clinically-grounded, interpretable survival predictions and improved subtyping. Specifically, the results on the ADNI cohort show that our framework incorporates unstructured clinical notes alongside structured neuroimaging data to guide interpretable expert routing, whereas traditional survival models rely on voxel-level correlations from the structured neuroimaging data. This is demonstrated by improved subtype separation over all baselines measured by LogRank, including a two-fold improvement over a standard MoE-based clustering framework, establishing iLENS as a strong baseline for interpretable AD conversion survival analysis. 

\section*{Limitations}
We also identify several limitations of the current framework, including reliance on LLM-based routing and expert initialization, as well as evaluation on a single cohort and disease setting. In this work, we focus on subtypes $K \in \{2,3\}$ for clinically interpretable low/high-risk stratification and moderate-granularity subtype discovery. Future work will assess robustness across different LLMs, survival outcomes, and subtyping settings.

\section*{Ethical Considerations}
Our proposed framework is intended for research and exploratory purposes on survival modeling. The generated expert assignments and natural language interpretability are based on LLM, which needs to be further evaluated for accuracy and hallucinations, especially when LLM-generated explanations may still appear clinically plausible even when uncertain or partially correct. At this stage, human clinical oversight remains necessary for any downstream clinical interpretation of patient assignments and rationales. 

\section*{Potential Risks}
Our proposed framework has several potential risks to be considered. First, the framework uses semantic routing that relies on LLM, which may inherit biases present in pretrained language models. Second, the generated expert descriptions also rely on semantic interpretations by the LLM on brain measurement features based on prior knowledge in the LLM’s training data. If prior knowledge contains incorrect information or LLM suffers from hallucination, it may lead to incorrect or unstable routing assignments that may disrupt prognosis and subtyping in certain cases. Third, the proposed framework was evaluated on ADNI cohort, which may not fully represent broader patient demographics, imaging protocols, or healthcare systems.

\bibliography{references}

@article{carobbio2020multistate,
  title={A multistate model of survival prediction and event monitoring in prefibrotic myelofibrosis},
  author={Carobbio, Alessandra and Guglielmelli, Paola and Rumi, Elisa and Cavalloni, Chiara and De Stefano, Valerio and Betti, Silvia and Rambaldi, Alessandro and Finazzi, Maria Chiara and Thiele, Juergen and Vannucchi, Alessandro M and others},
  journal={Blood cancer journal},
  volume={10},
  number={10},
  pages={100},
  year={2020},
  publisher={Nature Publishing Group UK London}
}

@article{glare2003systematic,
  title={A systematic review of physicians' survival predictions in terminally ill cancer patients},
  author={Glare, Paul and Virik, Kiran and Jones, Mark and Hudson, Malcolm and Eychmuller, Steffen and Simes, John and Christakis, Nicholas},
  journal={Bmj},
  volume={327},
  number={7408},
  pages={195},
  year={2003},
  publisher={British Medical Journal Publishing Group}
}

@article{binder2008allowing,
  title={Allowing for mandatory covariates in boosting estimation of sparse high-dimensional survival models},
  author={Binder, Harald and Schumacher, Martin},
  journal={BMC bioinformatics},
  volume={9},
  number={1},
  pages={14},
  year={2008},
  publisher={Springer}
}

@article{abbasi2024survival,
  title={Survival prediction landscape: an in-depth systematic literature review on activities, methods, tools, diseases, and databases},
  author={Abbasi, Ahtisham Fazeel and Asim, Muhammad Nabeel and Ahmed, Sheraz and Vollmer, Sebastian and Dengel, Andreas},
  journal={Frontiers in Artificial Intelligence},
  volume={7},
  pages={1428501},
  year={2024},
  publisher={Frontiers Media SA}
}

@article{katzman2018deepsurv,
  title={DeepSurv: personalized treatment recommender system using a Cox proportional hazards deep neural network},
  author={Katzman, Jared L and Shaham, Uri and Cloninger, Alexander and Bates, Jonathan and Jiang, Tingting and Kluger, Yuval},
  journal={BMC medical research methodology},
  volume={18},
  number={1},
  pages={24},
  year={2018},
  publisher={Springer}
}

@article{fox2002cox-ph,
  title={Cox proportional-hazards regression for survival data},
  author={Fox, John and Weisberg, Sanford},
  journal={An R and S-PLUS companion to applied regression},
  volume={2002},
  pages={7},
  year={2002}
}

@article{nagpal2021dsm,
  title={Deep survival machines: Fully parametric survival regression and representation learning for censored data with competing risks},
  author={Nagpal, Chirag and Li, Xinyu and Dubrawski, Artur},
  journal={IEEE Journal of Biomedical and Health Informatics},
  volume={25},
  number={8},
  pages={3163--3175},
  year={2021},
  publisher={IEEE}
}

@article{hou2024DCSM,
  title={Interpretable deep clustering survival machines for Alzheimer’s disease subtype discovery},
  author={Hou, Bojian and Wen, Zixuan and Bao, Jingxuan and Zhang, Richard and Tong, Boning and Yang, Shu and Wen, Junhao and Cui, Yuhan and Moore, Jason H and Saykin, Andrew J and others},
  journal={Medical image analysis},
  volume={97},
  pages={103231},
  year={2024},
  publisher={Elsevier}
}

@inproceedings{hou2023dcsm-ieee,
  title={Deep clustering survival machines with interpretable expert distributions},
  author={Hou, Bojian and Li, Hongming and others},
  booktitle={2023 IEEE 20th International Symposium on Biomedical Imaging (ISBI)},
  pages={1--4},
  year={2023},
  organization={IEEE}
}

@inproceedings{wen2025multimodel-dcsm,
  title={Multi-Modal Deep Clustering Survival Machines for Alzheimer's Disease Subtype Discovery},
  author={Wen, Zixuan and Hou, Bojian and others},
  booktitle={Proc. of the IEEE/CVF Int. Conf. on Computer Vision},
  pages={2264--2272},
  year={2025}
}

@inproceedings{jeanselme2022neural-nsc,
  title={Neural Survival Clustering: Non-parametric mixture of neural networks for survival clustering},
  author={Jeanselme, Vincent and Tom, Brian and Barrett, Jessica},
  booktitle={Conference on Health, Inference, and Learning},
  pages={92--102},
  year={2022},
  organization={PMLR}
}

@article{manduchi2021deep-vadesc,
  title={A deep variational approach to clustering survival data},
  author={Manduchi, Laura and Marcinkevi{\v{c}}s, Ri{\v{c}}ards and Massi, Michela C and Weikert, Thomas and Sauter, Alexander and Gotta, Verena and M{\"u}ller, Timothy and Vasella, Flavio and Neidert, Marian C and Pfister, Marc and others},
  journal={arXiv preprint arXiv:2106.05763},
  year={2021}
}

@inproceedings{chapfuwa2020survival-sca,
  title={Survival cluster analysis},
  author={Chapfuwa, Paidamoyo and Li, Chunyuan and Mehta, Nikhil and Carin, Lawrence and Henao, Ricardo},
  booktitle={Proceedings of the ACM Conference on Health, Inference, and Learning},
  pages={60--68},
  year={2020}
}

@article{fedus2022switch,
  title={Switch transformers: Scaling to trillion parameter models with simple and efficient sparsity},
  author={Fedus, William and Zoph, Barret and Shazeer, Noam},
  journal={Journal of Machine Learning Research},
  volume={23},
  number={120},
  pages={1--39},
  year={2022}
}

@article{xin2025i2moe,
  title={I2moe: Interpretable multimodal interaction-aware mixture-of-experts},
  author={Xin, Jiayi and Yun, Sukwon and Peng, Jie and Choi, Inyoung and Ballard, Jenna L and Chen, Tianlong and Long, Qi},
  journal={arXiv preprint arXiv:2505.19190},
  year={2025}
}

@article{zhuang2025mref,
  title={Multimodal Fusion of Regional Brain Experts for Interpretable Alzheimer's Disease Diagnosis},
  author={Zhuang, Farica and Aliyeva, Dinara and Yang, Shu and Wen, Zixuan and Duong-Tran, Duy and Davatzikos, Christos and Chen, Tianlong and Wang, Song and Shen, Li},
  journal={arXiv preprint arXiv:2512.10966},
  year={2025}
}

@article{yun2024flex,
  title={Flex-moe: Modeling arbitrary modality combination via the flexible mixture-of-experts},
  author={Yun, Sukwon and Choi, Inyoung and Peng, Jie and Wu, Yangfan and Bao, Jingxuan and Zhang, Qiyiwen and Xin, Jiayi and Long, Qi and Chen, Tianlong},
  journal={Advances in Neural Information Processing Systems},
  volume={37},
  pages={98782--98805},
  year={2024}
}

@article{zhang2025surmoe,
  title={Integrating images and genomics for multi-modal cancer survival analysis via mixture of experts},
  author={Zhang, Wei and Xu, Wenxin and Chen, Tong and Sakal, Collin and Li, Xinyue},
  journal={Information Fusion},
  pages={103521},
  year={2025},
  publisher={Elsevier}
}

@article{zhu2024unimed,
  title={Uni-med: a unified medical generalist foundation model for multi-task learning via connector-MoE},
  author={Zhu, Xun and Hu, Ying and Mo, Fanbin and Li, Miao and Wu, Ji},
  journal={Advances in Neural Information Processing Systems},
  volume={37},
  pages={81225--81256},
  year={2024}
}

@article{shen2024mome,
  title={Mome: Mixture of multimodal experts for generalist multimodal large language models},
  author={Shen, Leyang and Chen, Gongwei and Shao, Rui and Guan, Weili and Nie, Liqiang},
  journal={Advances in neural information processing systems},
  volume={37},
  pages={42048--42070},
  year={2024}
}

@inproceedings{hou2025adagent,
  title={Adagent: Llm agent for alzheimer’s disease analysis with collaborative coordinator},
  author={Hou, Wenlong and Yang, Guangqian and Du, Ye and Lau, Yeung and Liu, Lihao and He, Junjun and Long, Ling and Wang, Shujun},
  booktitle={International Workshop on Agentic AI for Medicine},
  pages={23--32},
  year={2025},
  organization={Springer}
}

@inproceedings{tang2024medagents,
  title={Medagents: Large language models as collaborators for zero-shot medical reasoning},
  author={Tang, Xiangru and Zou, Anni and Zhang, Zhuosheng and Li, Ziming and Zhao, Yilun and Zhang, Xingyao and Cohan, Arman and Gerstein, Mark},
  booktitle={Findings of the Association for Computational Linguistics: ACL 2024},
  pages={599--621},
  year={2024}
}

@article{wang2025medagent-pro,
  title={Medagent-pro: Towards evidence-based multi-modal medical diagnosis via reasoning agentic workflow},
  author={Wang, Ziyue and Wu, Junde and Cai, Linghan and Low, Chang Han and Yang, Xihong and Li, Qiaxuan and Jin, Yueming},
  journal={arXiv preprint arXiv:2503.18968},
  year={2025}
}

@inproceedings{li2024mmedagent,
  title={Mmedagent: Learning to use medical tools with multi-modal agent},
  author={Li, Binxu and Yan, Tiankai and Pan, Yuanting and Luo, Jie and Ji, Ruiyang and Ding, Jiayuan and Xu, Zhe and Liu, Shilong and Dong, Haoyu and Lin, Zihao and others},
  booktitle={Findings of the Association for Computational Linguistics: EMNLP 2024},
  pages={8745--8760},
  year={2024}
}

@article{kim2024mdagents,
  title={Mdagents: An adaptive collaboration of llms for medical decision-making},
  author={Kim, Yubin and Park, Chanwoo and Jeong, Hyewon and Chan, Yik S and Xu, Xuhai and McDuff, Daniel and Lee, Hyeonhoon and Ghassemi, Marzyeh and Breazeal, Cynthia and Park, Hae W},
  journal={Advances in Neural Information Processing Systems},
  volume={37},
  pages={79410--79452},
  year={2024}
}

@article{ong2024routellm,
  title={Routellm: Learning to route llms with preference data},
  author={Ong, Isaac and Almahairi, Amjad and Wu, Vincent and Chiang, Wei-Lin and Wu, Tianhao and Gonzalez, Joseph E and Kadous, M Waleed and Stoica, Ion},
  journal={arXiv preprint arXiv:2406.18665},
  year={2024}
}

@article{liu2025llmoe,
  title={LLM-Based Routing in Mixture of Experts: A Novel Framework for Trading},
  author={Liu, Kuan-Ming and Lo, Ming-Chih},
  journal={arXiv preprint arXiv:2501.09636},
  year={2025}
}

@article{hu2024routerbench,
  title={Routerbench: A benchmark for multi-llm routing system},
  author={Hu, Qitian Jason and Bieker, Jacob and Li, Xiuyu and Jiang, Nan and Keigwin, Benjamin and Ranganath, Gaurav and Keutzer, Kurt and Upadhyay, Shriyash Kaustubh},
  journal={arXiv preprint arXiv:2403.12031},
  year={2024}
}

@inproceedings{akiba2019optuna,
  title={Optuna: A next-generation hyperparameter optimization framework},
  author={Akiba, Takuya and Sano, Shotaro and Yanase, Toshihiko and Ohta, Takeru and Koyama, Masanori},
  booktitle={Proceedings of the 25th ACM SIGKDD international conference on knowledge discovery \& data mining},
  pages={2623--2631},
  year={2019}
}

@article{petersen2010alzheimer,
  title={Alzheimer's disease Neuroimaging Initiative (ADNI) clinical characterization},
  author={Petersen, Ronald Carl and others},
  journal={Neurology},
  volume={74},
  number={3},
  pages={201--209},
  year={2010},
  publisher={Lippincott Williams \& Wilkins}
}

@article{rolls2020-AAL3,
  title={Automated anatomical labelling atlas 3},
  author={Rolls, Edmund T and Huang, Chu-Chung and Lin, Ching-Po and Feng, Jianfeng and Joliot, Marc},
  journal={Neuroimage},
  volume={206},
  pages={116189},
  year={2020},
  publisher={Elsevier}
}

@article{ashburner2000-VBM,
  title={Voxel-based morphometry—the methods},
  author={Ashburner, John and Friston, Karl J},
  journal={Neuroimage},
  volume={11},
  number={6},
  pages={805--821},
  year={2000},
  publisher={Elsevier}
}

@article{nakagawa2020prediction,
  title={Prediction of conversion to Alzheimer’s disease using deep survival analysis of MRI images},
  author={Nakagawa, Tomonori and Ishida, Manabu and Naito, Junpei and Nagai, Atsushi and Yamaguchi, Shuhei and Onoda, Keiichi and Alzheimer’s Disease Neuroimaging Initiative},
  journal={Brain communications},
  volume={2},
  number={1},
  pages={fcaa057},
  year={2020},
  publisher={Oxford University Press}
}

@article{sperling2011toward,
  title={Toward defining the preclinical stages of Alzheimer’s disease: Recommendations from the National Institute on Aging-Alzheimer's Association workgroups on diagnostic guidelines for Alzheimer's disease},
  author={Sperling, Reisa A and Aisen, Paul S and Beckett, Laurel A and Bennett, David A and Craft, Suzanne and Fagan, Anne M and Iwatsubo, Takeshi and Jack Jr, Clifford R and Kaye, Jeffrey and Montine, Thomas J and others},
  journal={Alzheimer's \& dementia},
  volume={7},
  number={3},
  pages={280--292},
  year={2011},
  publisher={Elsevier}
}

@article{mirabnahrazam2023predicting,
  title={Predicting time-to-conversion for dementia of Alzheimer's type using multi-modal deep survival analysis},
  author={Mirabnahrazam, Ghazal and Ma, Da and Beaulac, Cedric and Lee, Sieun and Popuri, Karteek and Lee, Hyunwoo and Cao, Jiguo and Galvin, James E and Wang, Lei and Beg, Mirza Faisal and others},
  journal={Neurobiology of aging},
  volume={121},
  pages={139--156},
  year={2023},
  publisher={Elsevier}
}

@article{goyal2018characterizing,
  title={Characterizing heterogeneity in the progression of Alzheimer's disease using longitudinal clinical and neuroimaging biomarkers},
  author={Goyal, Devendra and Tjandra, Donna and Migrino, Raymond Q and Giordani, Bruno and Syed, Zeeshan and Wiens, Jenna and Alzheimer's Disease Neuroimaging Initiative and others},
  journal={Alzheimer's \& Dementia: Diagnosis, Assessment \& Disease Monitoring},
  volume={10},
  pages={629--637},
  year={2018},
  publisher={Elsevier}
}

\appendix

\section{Appendix}
\label{sec:appendix}

\subsection{Dataset Details}
\label{appendix:dataset}
For our experiment, we focus on 398 unique participants who were diagnosed with Mild Cognitive Impairment (MCI) and remained MCI in their last visit (censored) or converted to AD diagnosis at any point in the study (uncensored). We include their brain imaging measurements in tabular format mapped using Anatomical Automatic Labeling (AAL) \cite{rolls2020-AAL3}. Measurements include  structural magnetic resonance imaging (MRI), commonly used to detect atrophy in the brain, that were converted to Voxel-Based Morphometry (VBM) map \cite{ashburner2000-VBM}, as well as Florbetapir (AV45) and fluorodeoxyglucose (FDG) positron emission tomography (PET) imaging, which are used to detect the amyloid-$\beta$ (A$\beta$) load and brain glucose metabolism in the brain. Hence, the modalities are named VBM, FDG, and AV45 in the results table (Table~\ref{tab:multimodal_results}) of this work. This results in a table of quantitative measures for each region of interest (ROI) in the brain from the MRI image, where each value represents the relative gray matter volume within the specific brain structures. Additionally, we include summary descriptions from each participant's initial assessment and recent medical history. These are left as natural language. To construct the data for survival prediction, we labeled unique participants with whether the event did not occur (censored). For participants with multiple visits for brain imaging, we take the imaging information from the first MCI diagnosis. For patients who converted to AD, we take the date of the first AD diagnosis to calculate the number of days from the first MCI diagnosis. Using this data, we perform AD diagnosis for each participant into one of the three states. Data is split into 60/10/30 for training, validation, and test datasets, respectively.

The ADNI dataset was previously collected under institutional review board (IRB) approval and informed consent procedures. All experiments were conducted on de-identified research data. The ADNI data we use in this work is publicly available but requires approval for access. More information about the dataset and instructions on how to request for access are available through the ADNI website \url{https://adni.loni.usc.edu/}. We use the ADNI dataset consistently with its intended use for research on Alzheimer's disease.

\subsection{Training, Hyperparameter Tuning, and Evaluation}
\label{appendix:training_and_eval}
For all methods, we run Optuna \cite{akiba2019optuna} for hyperparameter tuning. We employ the Tree-structured Parzen Estimator (TPE) sampler for a total of 15 trials per model. This ensures a rigorous and fair comparison across all architectures by identifying optimal configurations for both baselines and our proposed framework. We maximize the validation metric for each C-index and LogRank across the number of clusters $K$ and imaging modalities (VBM, FDG, AV45). We report the resulting performance using the optimal hyperparameters on the test data in Table~\ref{tab:multimodal_results} based on the single data split. Optimal hyperparameters for iLENS are shown in Table~\ref{tab:hyperparameters}

\begin{table*}[t]
\centering
\small
\caption{Optimized hyperparameters for iLENS across modalities and latent subgroup settings. Separate Optuna studies were conducted for C-index and LogRank optimization objectives.}
\begin{tabular}{c c c c c c c}
\toprule
Modality & $K$ & Objective & LR & Discount & Dropout & Batch \\
\midrule

VBM & 2 & C-index & 4.76e-4 & 0.512 & 0.001 & 64 \\
VBM & 3 & C-index & 2.46e-3 & 0.980 & 0.155 & 64 \\

FDG & 2 & C-index & 3.10e-4 & 0.634 & 0.033 & 16 \\
FDG & 3 & C-index & 4.58e-5 & 0.811 & 0.062 & 32 \\

AV45 & 2 & C-index & 2.71e-3 & 0.588 & 0.152 & 16 \\
AV45 & 3 & C-index & 1.20e-3 & 0.967 & 0.198 & 16 \\

\midrule

VBM & 2 & LogRank & 3.71e-4 & 0.845 & 0.061 & 64 \\
VBM & 3 & LogRank & 1.14e-5 & 0.980 & 0.109 & 64 \\

FDG & 2 & LogRank & 2.52e-3 & 0.799 & 0.116 & 32 \\
FDG & 3 & LogRank & 4.81e-4 & 0.687 & 0.186 & 16 \\

AV45 & 2 & LogRank & 1.12e-5 & 0.574 & 0.135 & 16 \\
AV45 & 3 & LogRank & 1.66e-3 & 0.504 & 0.268 & 32 \\

\bottomrule
\end{tabular}

\label{tab:hyperparameters}

\end{table*}

\subsection{Subtyping}
\label{appendix:km_plots_av45_2_clusters}
Figure~\ref{fig:km-av45-k2} shows the Kaplan--Meier survival curves for AV45 PET amyloid imaging with $K=2$ latent survival subtypes across clustering survival methods. iLENS produces the clearest separation between high-risk and low-risk groups, resulting in the highest LogRank statistic reported in Table~\ref{tab:multimodal_results}. Other baseline methods show weaker separation or collapsed subtype assignments, such as for NSC where all samples are assigned to a single cluster, resulting in no subtyping. DCSM and VaDeSC are able to identify more distinguished survival curves, but they show less separation compared to iLENS. These results suggest that our LLM-guided expert routing framework is able to improve the discovery of AD progression patterns based on amyloid PET imaging.


















\subsection{Expert Design}
\label{appendix:expert_design}

In the first step of our LLM-routing framework, we performed expert design using the LLM. We prompted the LLM with modality-specific feature names, a description of the imaging modality and measurements, and representative clinical narrative fields, while constraining the output to four JSON-formatted expert phenotypes. The generated experts and descriptions allow the downstream LLM routing to interpret them in a consistent manner for the mixture-of-experts. To encourage clinically meaningful specialization of the experts, we explicitly constrained the LLM to generate experts corresponding to distinct progression mechanisms and multimodal phenotypes rather than feature missingness patterns or arbitrary partitions. The prompt additionally restricts the LLM to use only information available within the current modality. As a result, the generated experts differ across modalities (e.g., AV45 amyloid PET, FDG PET, and VBM MRI) and reflect modality-specific progression characteristics.

Table~\ref{tab:av45_prompt} shows an example the prompt template used for AV45 expert initialization, while Table~\ref{tab:av45_experts} shows the resulting  AV45 expert phenotypes generated by the LLM.

\begin{table*}[ht]
\centering

\begin{tcolorbox}[
    enhanced,
    width=\textwidth,
    title=LLM Expert Initialization Prompt,
    colback=gray!5,
    colframe=black!60,
    boxrule=0.5pt,
    arc=2mm,
    left=1mm,
    right=1mm,
    top=1mm,
    bottom=1mm,
]

\small

\textbf{System Prompt:}

You are an expert nuclear-medicine neurologist preparing phenotype experts for survival analysis.

Design experts for MCI-to-AD conversion risk using amyloid PET
(Florbetapir AV-45) SUVR mesoscale patterns plus clinical text.

You do not diagnose or recommend treatment.

Hard constraints:

- Use AV-45 regional SUVR-style signals (from av45\_* features)
  and clinical narratives (initial assessment / recent medical history).

- Do NOT assume FDG hypometabolism, structural MRI/VBM, CSF,
  or tau PET unless such columns appear in feature\_names.

- Experts must be clinically separable mechanisms
  (e.g. spatial amyloid distribution, comorbid context),
  not missingness buckets.

Output JSON only, no markdown.

Rules:

- num\_experts must be exactly 4.

- Expert ids must be 0..3.

\vspace{0.5em}

\textbf{Task Prompt:}

Propose 4 AD progression phenotype experts using AV-45 amyloid PET
(cortical SUVR) patterns plus initial assessment/recent medical
history narratives for subtype clustering and survival/log-rank separation.

\end{tcolorbox}

\caption{Prompt template used for AV45 expert initialization.}
\label{tab:av45_prompt}

\end{table*}

\begin{table*}[!ht]

\centering

\begin{tabular}{p{0.06\textwidth} p{0.18\textwidth} p{0.10\textwidth} p{0.56\textwidth}}
ID & Name & Risk & Description \\
\midrule
E0 & Vascular comorbidity low-amyloid phenotype & Low &
Low SUVR values across cortical regions (especially frontal, parietal, temporal) combined with clinical text indicating hypertension, stroke, or vascular disease. Highlights patients whose cognitive impairment may be driven more by vascular pathology than amyloid burden, explaining slower AD progression risk.\\
\midrule
E1 & Frontal amyloid with neuropsychiatric symptoms & Mid-High &
Elevated SUVR in frontal superior, middle, and inferior gyri plus clinical notes describing agitation, apathy, or executive dysfunction. Captures a phenotype with frontal lobe amyloid pathology linked to behavioral symptoms, associated with faster progression than low-amyloid vascular phenotype. \\
\midrule
E2 & Posterior cortical amyloid with memory impairment & Mid-High &
High SUVR in precuneus, posterior cingulate, parietal, and occipital cortices combined with clinical text emphasizing memory loss and cognitive decline. Represents the classical AD progression phenotype with posterior amyloid deposition and memory impairment, supporting established clinical-pathological correlations.\\
\midrule
E3 & Diffuse high amyloid with multiple comorbidities & High &
Elevated SUVR in frontal, parietal, temporal, and occipital regions combined with clinical notes indicating rapid progression, multiple medical issues, or neuropsychiatric plus memory symptoms. Identifies patients with extensive amyloid pathology and complex clinical profiles, associated with the highest risk of rapid MCI-to-AD conversion.\\
\bottomrule
\end{tabular}

\caption{LLM-initialized AV45 expert phenotypes used for semantic routing.}

\label{tab:av45_experts}

\end{table*}

\subsection{Mesoscale Brain Region Summarization}
\label{appendix:mesoscale}

The original ADNI imaging features consist of region-of-interest (ROI) measurements derived from the Automated Anatomical Labeling (AAL) atlas, where each anatomical structure is represented separately for the left and right hemispheres (e.g., \texttt{Frontal\_Sup\_L} and \texttt{Frontal\_Sup\_R}). There are a total of 116 ROI measurements in each modality. To reduce dimensionality and improve semantic interpretability for LLM routing, we aggregate anatomically related ROIs into mesoscale brain-region groups.

For each patient, we compute the mean across ROI-level SUVR measurements within each mesoscale group to produce compact regional summaries (e.g., Frontal-Lobe, Cingulate, Parietal-Lobe). These summaries are then provided to the LLM router as interpretable imaging descriptors alongside the clinical narrative summaries for each user input. The resulting mesoscale representation preserves broad spatial amyloid distribution patterns while substantially reducing prompt complexity. Table~\ref{tab:mesoscale_regions} shows representative mesoscale region groupings used.

\begin{table*}[t]
\centering
\small

\begin{tabular}{p{0.23\textwidth} p{0.70\textwidth}}
\toprule
Mesoscale Region & AAL ROIs Included \\
\midrule

Frontal-Lobe &
Frontal\_Sup, Frontal\_Sup\_Orb, Frontal\_Mid,
Frontal\_Mid\_Orb, Frontal\_Inf\_Oper,
Frontal\_Inf\_Tri, Frontal\_Inf\_Orb,
Rolandic\_Oper, Supp\_Motor\_Area,
Frontal\_Sup\_Medial, Frontal\_Med\_Orb,
Rectus \\

Insula &
Insula \\

Cingulate &
Cingulum\_Ant, Cingulum\_Mid,
Cingulum\_Post \\

Temporal-Lobe-Cortical &
Olfactory, ParaHippocampal,
Lingual, Fusiform,
Heschl, Temporal\_Sup,
Temporal\_Pole\_Sup,
Temporal\_Mid,
Temporal\_Pole\_Mid,
Temporal\_Inf \\

Temporal-Subcortical &
Hippocampus, Amygdala \\

Occipital-Lobe &
Calcarine, Cuneus,
Occipital\_Sup,
Occipital\_Mid,
Occipital\_Inf \\

Sensory-Motor-Cortex &
Precentral, Postcentral,
Paracentral\_Lobule \\

Parietal-Lobe &
Parietal\_Sup,
Parietal\_Inf,
SupraMarginal,
Angular,
Precuneus \\

Striatum/Basal-Ganglia &
Caudate, Putamen,
Pallidum \\

Thalamus &
Thalamus \\

Cerebellum &
Cerebelum\_Crus1,
Cerebelum\_Crus2,
Cerebelum\_3,
Cerebelum\_4\_5,
Cerebelum\_6,
Cerebelum\_7b,
Cerebelum\_8,
Cerebelum\_9,
Cerebelum\_10,
Vermis\_1\_2,
Vermis\_3,
Vermis\_4\_5,
Vermis\_6,
Vermis\_7,
Vermis\_8,
Vermis\_9,
Vermis\_10 \\

\bottomrule
\end{tabular}

\caption{Mesoscale region groupings derived from AAL atlas regions. Each anatomical structure contains both left and right hemisphere measurements, which are aggregated using mean SUVR summaries within each mesoscale category.}
\label{tab:mesoscale_regions}

\end{table*}

\subsection{LLM-routing}
\label{appendix:llm-routing}

During routing for the mixture-of-experts, the LLM receives only modality-specific imaging summaries across mesoscale brain regions, clinical text, and the previously generated expert definitions. We ensure that the router does not receive survival labels, conversion outcomes, censoring status, subtype assignments, or time-to-event information. We provide the prompt template used for AV45 routing in Table~\ref{tab:av45_llm_router_prompt} as an example. The prompt instructs the LLM to assign sparse mixture weights across phenotype experts using only imaging measurements and clinical notes while constraining routing to clinically meaningful progression patterns. We set TOP\_K to 2 for each modality to employ a sparse mixture-of-experts.

\begin{table*}[t]
\centering

\begin{tcolorbox}[
    enhanced,
    width=\textwidth,
    title=LLM Router Prompt Template for AV45 PET,
    colback=gray!5,
    colframe=black!60,
    boxrule=0.5pt,
    arc=2mm,
    left=1.5mm,
    right=1.5mm,
    top=1mm,
    bottom=1mm,
]

\small

\textbf{System Prompt:}

You are a routing controller for an Alzheimer disease (AD) mixture-of-experts survival model
(amyloid PET AV-45 + clinical text). Route patients to phenotype experts using imaging and history;
you do not diagnose, stage, or prescribe.

\vspace{0.4em}

\textbf{Imaging input:}
Mesoscale summaries from \texttt{av45\_*} columns (AAL atlas). Each value is the nanmean across
ROI-level SUVR columns within the bucket. These are per-patient regional means on the merged SUVR
scale in this run, not z-scores versus a cognitively normal reference.

\vspace{0.4em}

\textbf{Interpretation guide -- AV45:}
Values are SUVR-like cortical amyloid uptake. Higher regional values generally indicate greater
fibrillar amyloid deposition in that compartment, subject to acquisition and reference-region
normalization. Compare across mesoscale regions, such as precuneus/posterior cingulate, lateral
frontal, and temporal neocortex, to identify spatial amyloid burden patterns. Values near 0 may
reflect low uptake, reference choice, or sparse data; interpret alongside the clinical narrative.
Use as probabilistic phenotype routing for progression risk, not as a clinical decision.

\vspace{0.4em}

\textbf{Clinical text:}
\texttt{clinical\_narrative\_summary} is one paragraph. It explains \texttt{INITHEALTH} versus
\texttt{RECMHIST}, gives calendar-day offsets from the cohort anchor visit and AV45 PET acquisition,
and quotes the intake and recent-history narratives.

\vspace{0.4em}

Do not request FDG, MRI/VBM, CSF, or raw ROI tables beyond what is given.
Assign nonnegative mixture weights summing to 1 for at most \texttt{TOP\_K} experts.

\vspace{0.4em}

\textbf{Output format:}

\begin{verbatim}
{
  "weights": {"<expert_id>": <float>, ...},
  "rationale": "<string>"
}
\end{verbatim}

Use 1--\texttt{TOP\_K} nonzero experts and renormalize if needed.

\vspace{0.5em}

\textbf{Patient-specific input template:}

\begin{verbatim}
{
  "TOP_K": 2,
  "num_experts": 4,
  "expert_personas": [
    {
      "id": 0,
      "name": "Vascular comorbidity low-amyloid phenotype",
      "mechanism": "...",
      "focus": "...",
      "avoid_routing_when": "...",
      "expected_relative_risk": "lowest",
      "interpretability_value": "..."
    },
    { "id": 1, "...": "..." },
    { "id": 2, "...": "..." },
    { "id": 3, "...": "..." }
  ],
  "imaging_mesoscale_summary": {
    "summary_style": "regional_mean",
    "mesoscale_region_means": {
      "av45": {
        "Frontal-Lobe": 1.47,
        "Cingulate": 1.85,
        "...": "..."
      }
    }
  },
  "clinical_narrative_summary":
    "**Patient initial health assessment and medical history narrative** ..."
}
\end{verbatim}

\end{tcolorbox}

\caption{Prompt template used for AV45 PET semantic routing. The router receives imaging summaries, clinical narrative summaries, and expert definitions, but not survival outcomes or subtype labels.}
\label{tab:av45_llm_router_prompt}

\end{table*}

\subsection{LLM-guided Survival Modeling}
\label{appendix:survival-model-after-llm-routing}
We train a survival clustering framework using the LLM-routed sparse mixture-of-experts (MoE) representation layer. The experts are initialized as identical neural networks. During training, we obtain the LLM-assigned weights during LLM routing. The assigned experts then receive only structured modality-specific imaging features as numeric inputs without information from clinical notes. Clinical notes instead influences the model indirectly through the precomputed LLM routing weights generated during previous LLM routing stage.

More specifically, the representation layer consists of $E=4$ parallel expert multilayer perceptrons (MLPs), where each expert maps the imaging input $x$ into a latent embedding $h_e(x) \in \mathbb{R}^d$. The final representation is constructed as a weighted combination of expert embeddings $\Phi(x) = \sum_{e=1}^{E} w_e h_e(x)$, where $\mathbf{w}$ denotes the sparse LLM-generated routing weights. Unlike conventional MoE architectures with learnable gating networks, the routing weights are frozen during optimization and retrieved from precomputed LLM routing outputs. Only the expert subnetworks and downstream survival parameters are updated during training.

After the MoE representation layer, the model learns a latent mixture of $K$ Weibull survival components, where $K \in \{2,3\}$ in our experiments. These latent components define statistical survival subgroups used for survival prediction and subtype discovery \cite{hou2024DCSM}. 

At inference time, the fitted Weibull mixture model produces individualized survival risk estimates. For Kaplan--Meier and LogRank evaluation, patients are assigned to latent survival subgroups based on the highest predicted subgroup probability, and clusters are ordered from low to high risk according to mean predicted survival risk.

\subsection{Qualitative Routing Examples}
\label{appendix:interpretability}
To further examine the interpretability of our proposed framework, we provide representative example of LLM-guided semantic routing using AV45 amyloid PET imaging and clinical narrative summaries. During routing, the LLM receives only modality-specific imaging summaries, clinical text, and the previously generated expert definitions as presented in the previous subsection (Appendix~\ref{appendix:llm-routing}). The prompt format for the LLM router is provided in Table~\ref{tab:av45_llm_router_prompt}

Table~\ref{tab:semantic_routing_example} presents the output example from a representative patient who later converted from MCI to AD. The example shows how the LLM combines spatial amyloid distribution patterns with clinical comorbidity information to produce semantically interpretable expert assignments. In this case, the router assigns the highest weight to the diffuse high-amyloid phenotype while also partially routing the patient to the posterior cortical amyloid phenotype, reflecting mixed progression characteristics observed from the imaging and clinical narrative context.

\begin{table*}[t]
\centering

\begin{tcolorbox}[
    enhanced,
    width=\textwidth,
    title=Semantic Routing Output Example (AV45 PET),
    colback=gray!5,
    colframe=black!60,
    boxrule=0.5pt,
    arc=2mm,
    left=1mm,
    right=1mm,
    top=1mm,
    bottom=1mm,
]

\small

\textbf{Patient Imaging Summary:}
Diffuse elevated AV-45 SUVR values across frontal (1.47),
cingulate (1.85), temporal (1.35), parietal (1.39),
occipital (1.32), striatal (1.56), and thalamic (1.53) regions.

\vspace{0.4em}

\textbf{Clinical Narrative Summary:}
History includes anxiety (citalopram), sleep apnea (CPAP),
hypothyroidism, and hyperlipidemia.

\vspace{0.4em}

\textbf{LLM Routing Weights:}

E2 (Posterior cortical amyloid): 0.30 \\
E3 (Diffuse high amyloid): 0.70

\vspace{0.4em}

\textbf{LLM Rationale:}

The patient shows diffuse high amyloid SUVR values across multiple cortical regions including frontal (1.47), cingulate (1.85), temporal (1.35), parietal (1.39), occipital (1.32), and subcortical areas (striatum 1.56, thalamus 1.53). This widespread elevated amyloid burden aligns with the ``Diffuse high amyloid with multiple comorbidities'' phenotype (expert 3). Clinical text indicates multiple comorbidities such as mild anxiety, sleep apnea, hypothyroidism, and hyperlipidemia, supporting routing to expert 3. A secondary weight is assigned to expert 2. Overall, the patient fits best with expert 3 (highest risk) and to a lesser extent expert 2 (posterior cortical amyloid with memory impairment).

\vspace{0.4em}

\textbf{Subtype Cluster:} High risk

\vspace{0.4em}

\textbf{Outcome (hidden from router):}
Converted to AD after 723 days.

\end{tcolorbox}

\caption{Representative semantic routing output example from the AV45 PET modality. The router receives only imaging summaries, clinical narrative text, and expert definitions, without access to survival labels or conversion outcomes.}

\label{tab:semantic_routing_example}

\end{table*}

\subsection{Routing Ablation Studies}
\label{appendix:ablations}

We performed additional ablation experiments to evaluate the contribution of semantic LLM-guided routing compared to non-semantic or non-specialized routing strategies. Specifically, we compared against (i) uniform dense routing, where all experts receive equal weights, (ii) uniform sparse routing with equal sparse weights across selected experts, (iii) random sparse routing, and (iv) LLM routing without clinical narrative information, where routing is generated using imaging summaries only.

Tables~\ref{tab:ablation_av45}, \ref{tab:ablation_fdg}, and \ref{tab:ablation_vbm} summarize the results across AV45 PET, FDG PET, and VBM MRI modalities for both $K=2$ and $K=3$ latent survival subgroup settings.

Overall, semantic LLM-guided routing consistently outperformed sparse non-semantic routing variants, including random sparse routing and uniform sparse routing, particularly for LogRank to measure subtype separation. The strongest improvements can be seen in AV45 PET and FDG PET, where semantic routing substantially improved survival subgroup stratification. Removing clinical narrative information generally reduced LogRank performance, suggesting that clinical text contributes information beyond imaging measurements alone.

Interestingly, uniform dense routing remains competitive in several ablation settings, indicating that averaging across all experts can provide stable representations even without sparse semantic specialization. However, we note that the dense setting increases model complexity. However, one downside with dense routing setting is that it removes patient-specific expert selection and does not provide interpretable semantic routing assignments. In contrast, our proposed LLM-guided routing framework produces sparse, clinically interpretable expert allocations while achieving competitive or improved survival stratification performance across multiple modality and subgroup configurations.

\begin{table*}[t]
\centering
\small

\begin{tabular}{lcccc}
\toprule
\multirow{2}{*}{Routing Variant} &
\multicolumn{2}{c}{$K=2$} &
\multicolumn{2}{c}{$K=3$} \\
\cmidrule(lr){2-3}
\cmidrule(lr){4-5}
&
C-index $\uparrow$ &
LogRank $\uparrow$ &
C-index $\uparrow$ &
LogRank $\uparrow$ \\
\midrule

Uniform Dense &
0.7828 & 0.0000 &
0.7826 & 43.2742 \\

Uniform Sparse &
0.7397 & 3.7351 &
0.7708 & 8.4878 \\

Random Sparse &
0.7087 & 1.1750 &
0.7501 & 14.5708 \\

LLM Routing w/o Clinical Notes &
0.7810 & 23.9817 &
0.7842 & 40.9760 \\

iLENS &
0.7590 & 36.3400 &
0.7670 & 29.0100 \\

\bottomrule
\end{tabular}

\caption{Routing ablation study for AV45 PET imaging.}
\label{tab:ablation_av45}

\end{table*}

\begin{table*}[t]
\centering
\small

\begin{tabular}{lcccc}
\toprule
\multirow{2}{*}{Routing Variant} &
\multicolumn{2}{c}{$K=2$} &
\multicolumn{2}{c}{$K=3$} \\
\cmidrule(lr){2-3}
\cmidrule(lr){4-5}
&
C-index $\uparrow$ &
LogRank $\uparrow$ &
C-index $\uparrow$ &
LogRank $\uparrow$ \\
\midrule

Uniform Dense &
0.7695 & 21.6903 &
0.7609 & 13.2011 \\

Uniform Sparse &
0.7396 & 0.5550 &
0.5133 & 2.3439 \\

Random Sparse &
0.6801 & 1.5189 &
0.6108 & 0.3452 \\

LLM Routing w/o Clinical Notes &
0.7806 & 18.0632 &
0.7488 & 1.9295 \\

iLENS &
0.7750 & 11.2200 &
0.7220 & 25.9700 \\

\bottomrule
\end{tabular}

\caption{Routing ablation study for FDG PET imaging.}
\label{tab:ablation_fdg}

\end{table*}

\begin{table*}[t]
\centering
\small

\begin{tabular}{lcccc}
\toprule
\multirow{2}{*}{Routing Variant} &
\multicolumn{2}{c}{$K=2$} &
\multicolumn{2}{c}{$K=3$} \\
\cmidrule(lr){2-3}
\cmidrule(lr){4-5}
&
C-index $\uparrow$ &
LogRank $\uparrow$ &
C-index $\uparrow$ &
LogRank $\uparrow$ \\
\midrule

Uniform Dense &
0.6667 & 1.8322 &
0.6844 & 10.3793 \\

Uniform Sparse &
0.6562 & 0.2236 &
0.6420 & 0.9036 \\

Random Sparse &
0.5890 & 0.4313 &
0.5662 & 2.7410 \\

LLM Routing w/o Clinical Notes &
0.6700 & 0.0048 &
0.6631 & 1.8550 \\

iLENS &
0.6700 & 2.9200 &
0.6340 & 2.5400 \\

\bottomrule
\end{tabular}

\caption{Routing ablation study for VBM MRI imaging.}
\label{tab:ablation_vbm}

\end{table*}

\subsection{AI Assistance in Research}
We used AI assistance (e.g., Cursor, ChatGPT) to support grammatical checks and writing revisions, coding frameworks and prompt design, and visualization generation.

\subsection{Computational Resources.}
Experiments were conducted on Apple Silicon hardware using an Apple M4 processor with 16GB unified memory.

\end{document}